\documentclass[10pt,twocolumn,letterpaper]{article}

\usepackage[pagenumbers]{cvpr} 
\usepackage{multirow}
\usepackage{booktabs}
\usepackage{color}
\usepackage{amssymb}
\usepackage{bm}
\usepackage{graphicx}
\usepackage{array}
\usepackage{indentfirst}
\usepackage{amsmath}
\usepackage{colortbl} 

\definecolor{lightblue}{rgb}{0.78, 0.082, 0.51}

\usepackage[dvipsnames]{xcolor}

\definecolor{cvprblue}{rgb}{0.21,0.49,0.74}
\usepackage[pagebackref,breaklinks,colorlinks,citecolor=cvprblue]{hyperref}

\def\paperID{8196} 
\def\confName{CVPR}
\def\confYear{2024}

\title{3DSFLabelling: Boosting 3D Scene Flow Estimation by Pseudo Auto-labelling}

\author{Chaokang Jiang$^1$, Guangming Wang$^2$, Jiuming Liu$^3$, Hesheng Wang$^3$, Zhuang Ma$^1$, \\ Zhenqiang Liu$^1$, Zhujin Liang$^1$, Yi Shan$^1$, Dalong Du$^{1,\dag
}$
\\
$^1$PhiGent Robotics, $^2$University of Cambridge, $^3$Shanghai Jiaotong University \\
{\tt\small ts20060079a31@cumt.edu.cn}, {\tt\small gw462@cam.ac.uk},
{\tt\small \{liujiuming, wanghesheng\}@sjtu.edu.cn}, \\
{\tt\small mazhuang097@outlook.com}, 
{\tt\small \{zhenqiang.liu, zhujin.liang, yi.shan, long.du\}@phigent.ai} \\
{\tt\small \href{https://github.com/jiangchaokang/3DSFLabelling}{github.com/jiangchaokang/3DSFLabelling} }
}

\begin{document}
\maketitle
\begin{abstract}
Learning 3D scene flow from LiDAR point clouds presents significant difficulties, including poor generalization from synthetic datasets to real scenes, scarcity of real-world 3D labels, and poor performance on real sparse LiDAR point clouds. We present a novel approach from the perspective of auto-labelling, aiming to generate a large number of 3D scene flow pseudo labels for real-world LiDAR point clouds. Specifically, we employ the assumption of rigid body motion to simulate potential object-level rigid movements in autonomous driving scenarios. By updating different motion attributes for multiple anchor boxes, the rigid motion decomposition is obtained for the whole scene. Furthermore, we developed a novel 3D scene flow data augmentation method for global and local motion. By perfectly synthesizing target point clouds based on augmented motion parameters, we easily obtain lots of 3D scene flow labels in point clouds highly consistent with real scenarios. On multiple real-world datasets including LiDAR KITTI, nuScenes, and Argoverse, our method outperforms all previous supervised and unsupervised methods without requiring manual labelling. Impressively, our method achieves a tenfold reduction in EPE3D metric on the LiDAR KITTI dataset, reducing it from $0.190m$ to a mere $0.008m$ error.
\end{abstract}

\section{Introduction}
\label{sec:intro}
\begin{figure}
  \centering
  \includegraphics[width=0.475\textwidth]{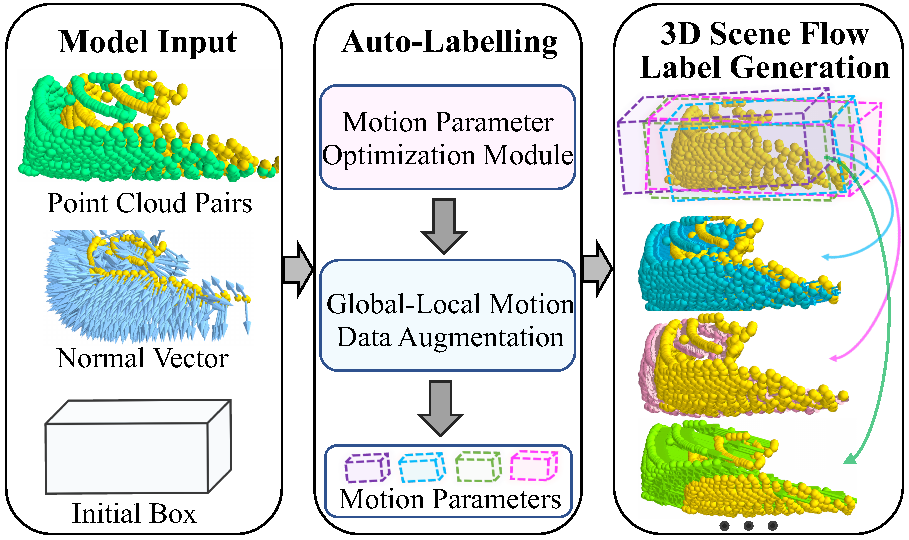}
  \vspace{-6mm}
  \caption{The proposed 3D scene flow pseudo-auto-labelling framework. Given point clouds and initial bounding boxes, both global and local motion parameters are iteratively optimized. Diverse motion patterns are augmented by randomly adjusting these motion parameters, thereby creating a diverse and realistic set of motion labels for the training of 3D scene flow estimation models.}
  \vspace{-2mm}
  \label{fig:intro}

\end{figure}

3D scene flow estimation through deducing per-point motion filed from consecutive frames of point clouds, serves a critical role across various applications, encompassing motion prediction \cite{wu2020motionnet, najibi2022motion}, anomaly motion detection \cite{iqbal2021detection}, 3D object detection \cite{erccelik20223d,zhang2020sdp}, and dynamic point cloud accumulation \cite{huang2022dynamic}. With the advancing of deep learning on point clouds \cite{qi2017pointnet,qi2017pointnet++}, many works \cite{Flownet3d,Flot,fu2023pt,Shen_2023_CVPR,jin2022deformation,cheng2023multi,zhang2023gmsf} have developed the learning-based methods to estimate per-point motion from 3D point clouds. Some state-of-the-art methods \cite{Shen_2023_CVPR,cheng2023multi,zhang2023gmsf} have reduced the average 3D EndPoint Error (EPE3D) to a few centimetres on the KITTI Scene Flow dataset (stereoKITTI) \cite{sfkitti,sfkitti2}. However, due to the scarcity of scene flow labels, these methods rely heavily on synthetic datasets such as FlyingThings3D (FT3D) \cite{FT3D} for network training. 


When evaluated on the stereoKITTI dataset \cite{sfkitti,sfkitti2}, PV-RAFT \cite{PVRAFT} demonstrates an average EPE3D of just $0.056m$. However, when evaluated on the Argoverse dataset \cite{argoverse}, the EPE3D metric astonishingly exceeds $10m$ \cite{FastNSF}. Therefore, learning 3D scene flow on synthetic dataset \cite{FT3D} has a large gap with real-world application. Jin et al. \cite{jin2022deformation} recently introduce a new synthetic dataset, GTA-SF, simulating LiDAR scans for autonomous driving. They propose a teacher-student domain adaptation framework to reduce the gap between synthetic and real datasets and improve some performance of 3D scene flow estimation. However, their performance is still poor in real-world LiDAR data because of ideal sensor models and lack of scene variety. Ideally, models should learn from real sensor data in the autonomous driving field. However, labelling each point's 3D motion vector for the 3D scene flow task is extremely costly. This has driven many works \cite{ding2022self,Justgo,Shen_2023_CVPR,li2022rigidflow,Flownet3d,wang2022sfgan} towards unsupervised or self-supervised learning of 3D scene flow. Although these methods have achieved reasonable accuracy, they still fall behind supervised methods, highlighting the importance of real sensor data and corresponding 3D scene flow labels. 

\begin{figure}
  \centering
  \includegraphics[width=0.47\textwidth]{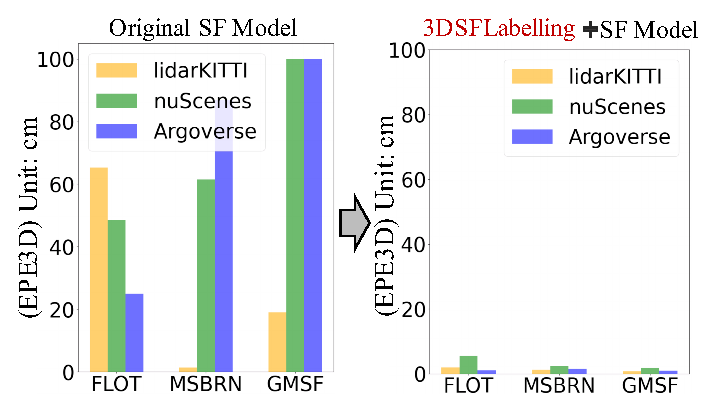}
  \vspace{-2mm}
  \caption{The accuracy improvement after integrating our proposed pseudo-auto-labelling method. Models trained on synthetic data performance poorly in 3D scene flow estimation for LiDAR-based autonomous driving. Our proposed 3D pseudo-auto-labelling method improves accuracy, reaching an EPE3D below $2cm$ across datasets \cite{sfkitti2,argoverse,nuscenes}.}
  \label{fig:intro2}
\end{figure}
In this work, we address three key challenges in the field of autonomous driving: the reliance on synthetic datasets that still have a poor generalization with real-world scenarios, the scarcity of scene flow labels in actual driving scenes, and the poor performance of existing 3D scene flow estimation networks on real LiDAR data. Inspired by the rigid motion assumptions in RigidFlow \cite{li2022rigidflow} and RSF \cite{deng2023rsf}, we propose a novel scene flow auto-labelling approach that leverages the characteristics of rigid motion prevalent in autonomous driving scenarios (Fig. \ref{fig:intro}). Specifically, we utilize 3D anchor boxes to segment 3D objects in point clouds. The attributes of each object-level box are not only position and size but also rotation, translation, motion status, and normal vector attributes. By leveraging the constrained loss functions for the box parameters and inter-frame association, we optimize the attributes of the boxes, subsequently combining these parameters with the source point cloud to produce a realistic target point cloud. Importantly, the generated target point cloud maintains a one-to-one correspondence with the source point cloud, enabling the efficient generation of pseudo 3D scene flow labels. 

To capture a more diverse range of motion patterns, we introduce a novel data augmentation strategy for 3D scene flow auto-labelling. Utilizing the attributes of each box, we simulate the rotations, translations, and motion status of both the ego vehicle and surrounding environment by adding Gaussian noise to these attributes. Consequently, we obtain numerous 3D scene flow labels with diverse motions that closely resemble real-world scenarios, furnishing the neural network with rich real training data and significantly improving the generalization capabilities of learning-based methods. Experimental results validate that our pseudo-label generation strategy consistently achieves state-of-the-art scene flow estimation results across various models \cite{Flot,cheng2023multi,zhang2023gmsf} and datasets \cite{sfkitti,argoverse,nuscenes} (Fig. \ref{fig:intro2}).

In summary, our contributions are as follows:
\begin{itemize} 
\item We propose a new framework for the automatic labelling of 3D scene flow pseudo-labels, significantly enhancing the accuracy of current scene flow estimation models, and effectively addressing the scarcity of 3D flow labels in autonomous driving.
\item We propose a universal 3D box optimization method with multiple motion attributes. Building upon this, we further introduce a plug-and-play 3D scene flow augmentation module with global-local motions and motion status. This allows for flexible motion adjustment of ego-motion and dynamic environments, setting a new benchmark for scene flow data augmentation.
\item Our method achieves state-of-the-art performance on KITTI, nuScenes, and Argoverse LiDAR datasets. Impressively, our approach surpasses all supervised and unsupervised methods without requiring any synthesising data and manual scene flow labels.
\end{itemize}


\section{Related Work}
\label{sec:RelatedWork}

\subsection{Supervised 3D Scene Flow Learning}
In recent years, the performance of methods \cite{Flownet3d, peng2023delflow,wang2021hierarchical} for 3D scene flow based on point cloud deep learning has surpassed traditional methods. FlowNet3D \cite{Flownet3d} pioneers an end-to-end approach to learning 3D scene flow from point clouds. Some works, such as HALFlow \cite{Hplflownet}, 3DFlow \cite{wang2022matters}, PointPWC \cite{Pointpwc}, and WSAFlowNet \cite{wang2023exploiting}, utilize PWC structures to learn 3D scene flow in a coarse-to-fine manner. Other methods address the disorderliness of points by voxelizing point clouds and using sparse convolution or voxel correlation fields to learn 3D scene flow, such as PV-RAFT \cite{PVRAFT}, DPV-RAFT \cite{DPVRAFT}, and SCTN \cite{li2022sctn}. Additional work refines the estimated scene flow through iterative procedures. MSBRN \cite{cheng2023multi} proposes bidirectional gated recurrent units for iteratively estimating scene flow. GMSF \cite{zhang2023gmsf} and PT-FlowNet \cite{fu2023pt} introduce point cloud transformers into 3D scene flow estimation networks. These supervised learning methods for 3D scene flow heavily rely on ground truth and are all trained on the FT3D dataset \cite{FT3D} and evaluated on stereoKITTI \cite{sfkitti,sfkitti2} for network generalization test.

\subsection{Unsupervised 3D Scene Flow Learning}
JGwF \cite{Justgo} and PointPWC \cite{Pointpwc} initially propose several self-supervised learning losses such as cycle consistency loss and chamfer loss. EgoFlow \cite{tishchenko2020self} distinguishes 3D scene flow into ego-motion flow and remaining non-rigid flow, achieving self-supervised learning based on temporal consistency. SFGAN \cite{wang2022sfgan} introduces generative adversarial concepts into self-supervised learning for 3D scene flow. Recently, works like R3DSF \cite{gojcic2021weakly}, RigidFlow \cite{li2022rigidflow}, and LiDARSceneFlow \cite{dong2022exploiting} greatly improve the accuracy of 3D scene flow estimation by introducing local or object-level rigidity constraints. RigidFlow \cite{li2022rigidflow} explicitly enforces rigid alignment within super-voxel regions by decomposing the source point cloud into multiple super-voxels. R3DSF \cite{gojcic2021weakly} separately considers background and foreground object-level 3D scene flow, relying on segmentation and odometry tasks.

\subsection{3D Scene Flow Optimization}
3D scene flow optimization techniques have demonstrated remarkable generalization capabilities, attracting a significant amount of academic research recently. Graph prior \cite{Pontes} optimizes scene flow to be as smooth as possible by using the Laplacian of point clouds. Some techniques introduce neural networks to optimize 3D scene flow. NSFP \cite{li2021neural} introduces a novel implicit regularizer, the Neural Scene Flow Prior, which primarily depends on runtime optimization and robust regularization. RSF \cite{deng2023rsf} combines global ego-motion with object-specific rigid movements to optimize 3D bounding box parameters and compute scene flow. FastNSF \cite{FastNSF} also adopts neural scene flow prior, and it shows more advantages in dealing with dense LiDAR points compared to learning methods. SCOOP \cite{lang2023scoop}, in the runtime phase, directly optimizes the flow refinement module using self-supervised objectives. Although optimization-based approaches for 3D scene flow estimation have demonstrated impressive accuracy, they typically involve high computational costs. 
\section{3DSFLabelling}
\label{sec:method}
3D scene flow estimation infers the 3D flow, $SF_{pred} \in \mathbb{R}^{3\times N_1}$ from the source point cloud $PC_{S} \in \mathbb{R}^{3\times N_1}$ and the target point cloud $PC_{T}\in \mathbb{R}^{3\times N_2}$ for each point in the source point. 
Previous self-supervised learning methods \cite{Pointpwc,Justgo} typically use the estimated 3D motion vector $SF_{pred}$ to warp the source point cloud $PC_S$ to the target point cloud $PC_{Sw}$. By comparing the difference between $PC_{Sw}$ and $PC_{T}$, a supervisory signal is generated. 
 
 In contrast with previous self-supervised learning methods, we propose bounding box element optimization to obtain the boxes and the box motion parameters from raw unlabelled point cloud data. Then, we use object-box-level motion parameters and global motion parameters to warp each box's points and the whole point cloud to the target point cloud, generating corresponding pseudo 3D scene flow labels. During the warping process of each object box, we propose augmenting the motion attributes of each object and the whole scene. This diversity assists the network in capturing a broader range of motion behaviours. 
\begin{figure*}[t]
  \centering
  \includegraphics[width=0.97\textwidth]{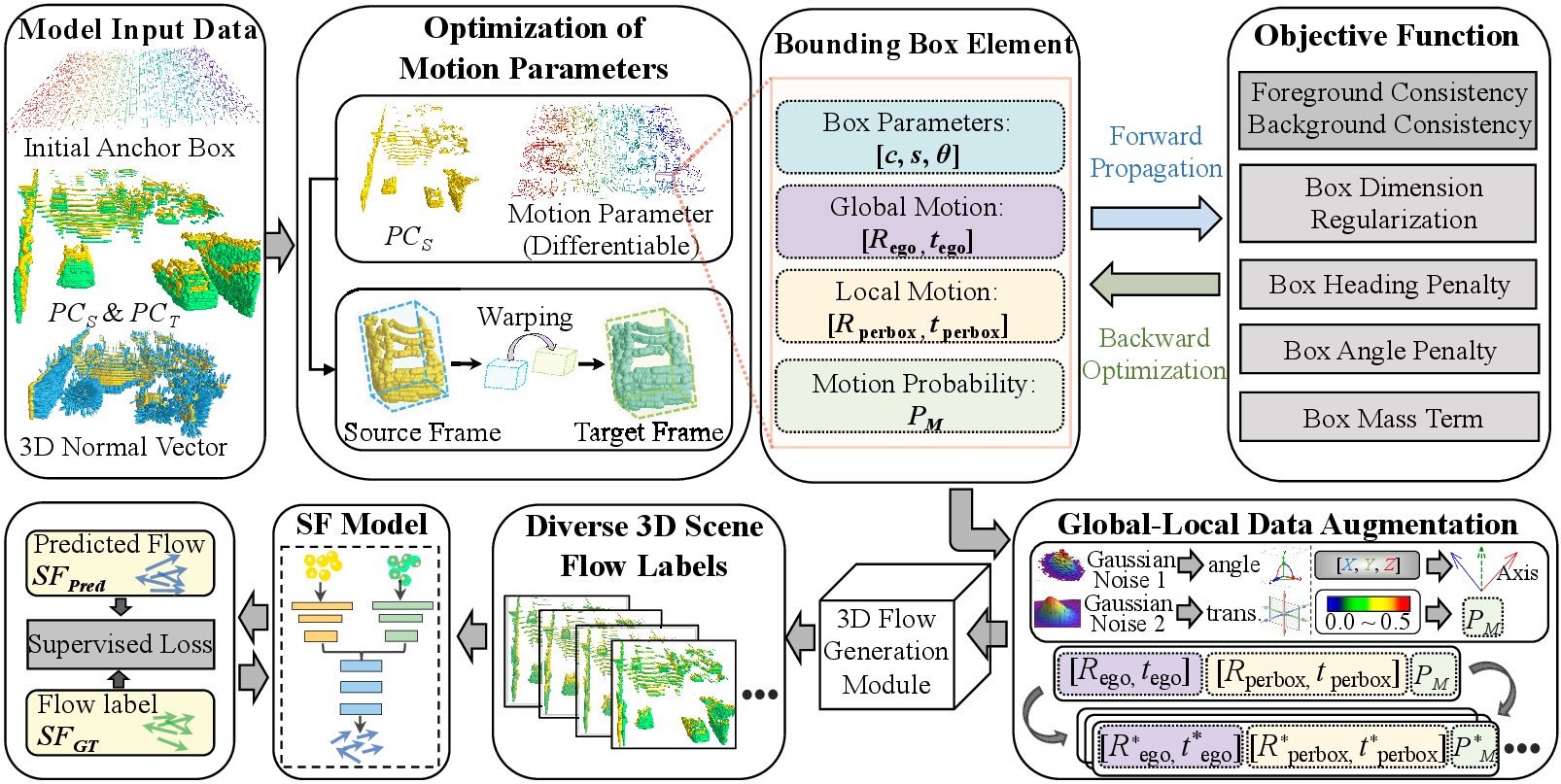}
  \vspace{-3mm}
  \caption{The proposed learning framework of pseudo 3D scene flow automatic labelling. The input comprises 3D anchor boxes, a pair of point clouds, and their corresponding coarse normal vectors. The optimization of motion parameters primarily updates the bounding box parameters, global motion parameters, local motion parameters, and the motion probability of the boxes. The attribute parameters for boxes are updated through backward optimization from six objective functions. Once optimized, the motion parameters simulate various types of motion using a global-local data augmentation module. A single source frame point cloud, along with the augmented motion parameters, produces diverse 3D scene flow labels. These labels serve to guide the supervised neural network to learn point-wise motion.}
  \label{fig:method}
\end{figure*}

\subsection{Prerequisites}
Apart from the two input point clouds, we do not require any extra labels, such as object-level tracking and semantic information, or vehicle ego-motion labels. To reinforce the geometric constraints in the pseudo label generation module, we employ Open3d \cite{zhou2018open3d} to generate coarse per-point normals. Despite these normals not being perfectly accurate, they are readily obtainable and can provide useful geometric constraints. Finally, we establish initial 3D anchor boxes with specific centers ($x, y, z$), width $w$, length $l$, height $h$, and rotation angle $\theta$, in accordance with the range of input points. As depicted in Fig. \ref{fig:method}, the inputs of our model consist of the initial anchor box set, $PC_S$, $PC_T$, and point cloud normals $N_S$.

\subsection{Motion Parameter Optimization Module}
As shown in Fig. \ref{fig:method}, we present the process of simulating the motion of point clouds in actual autonomous driving by updating four sets of parameters: differentiable bounding boxes $\Phi = [c, s, \theta]$, global motion parameters $\Theta = [R_{ego}, t_{ego}]$, motion parameters for each box $[R_{perbox}, t_{perbox}]$, and motion probability $P_M$ for each box. The variables \( c \), \( s \), and \( \theta \) represent the center coordinates, size, and orientation of the 3D box, respectively.

Inspired by RSF \cite{deng2023rsf}, 
we use the motion of object-level bounding boxes to present the point-wise 3D motion and make the step-like boxes differentiable through sigmoid approximation. By transforming the individual points to the bounding boxes, we introduce an object-level perception of the scene, enabling a more natural capture of rigid motion. This method proves advantageous in autonomous driving scenarios, where most objects predominantly exhibit rigid behaviour \cite{gojcic2021weakly}. Additionally, in the context of autonomous driving, most scene motion is typically caused by the ego motion of the vehicle. Hence, setting global motion parameters is necessary to simulate the global consistent rigid motion of the whole scene. To discern whether the motion of each box is caused by ego-motion, we also set up a motion probability for each bounding box. 

With the initial set of four motion parameters, the source point cloud is warped to the target frame, as follows:
\begin{equation}
\label{eq:1}
PC_{T}^{\Theta}, PC_{T}^{\Phi} = \Omega_1(\Theta, PC_S), \Omega_2(\Upsilon(\Phi, PC_S)),
\end{equation}
where $\Theta$ represents global motion parameters. $\Phi$ represents motion parameters of each bounding box, and $\Omega_1$ and $\Omega_2$ are background and foreground warping functions, respectively, generating the warped point clouds $PC_{T}^{\Theta}$ and $PC_{T}^{\Phi}$. $\Upsilon$ signifies the removal of boxes with too few points.

Based on the real target frame of point cloud and the generated target point clouds $PC_{T}^{\Theta}$ and $PC_{T}^{\Phi}$, we define loss functions to update and optimize the box attributes. We separately calculate the background and foreground losses:
\begin{equation}
\label{eq:2}
L_{BG} = \kappa(N_{T}^{\Theta} \oplus PC_{T}^{\Theta}, N_{T} \oplus PC_{T}) + \delta(PC_{T}^{\Theta}, PC_{T}),
\end{equation}
\begin{equation}
\label{eq:3}
\begin{aligned}
L_{FG} = \frac{1}{K_{box}} \sum P_M \times (\kappa(N_{T}^{\Phi} \oplus PC_{T}^{\Phi}, N_{T} \oplus PC_{T}) \\ + \delta(PC_{T}^{\Phi}, PC_{T})),
\end{aligned}
\end{equation}
where $\kappa$ is a function calculating nearest neighbour matches between the transformed point cloud and the target point cloud. $\delta$ is a pairwise distance function with location encoding. $K_{box}$ is the number of boxes, $P_M$ is the motion probability of each box, and the term $N_{T}\oplus PC_{T}$ represents the concatenation of the target point cloud's normal and positions.
As for the motion probability $P_M$ of each box:
\begin{equation}
\label{eq:4}
P_M = \sigma(\alpha \times (\Omega_3(\Phi,\gamma_i) + \beta_i)) - \alpha \times (\Omega_3(\Phi,\gamma_i) - \beta_i)),
\end{equation}
where $\sigma(x)$ represents the sigmoid function, $\alpha$ is a hyper-parameter `slope' in the sigmoid, $\beta$ represents the half size of the vector of 3D dimensions $w$, $l$, and $h$ of the bounding box. Coordinate values $\gamma$ in the source point cloud are warped to the target point cloud via motion box parameters $\Phi$. For each dynamic box, each point's relative position to the box's centre is calculated. Higher motion probability $P_M$ is assigned to the points closer to the centre. A fixed hyperparameter $\alpha$, controlling motion probability, may not effectively respond to diverse and complex autonomous driving scenarios. Therefore, we adopt an adaptive computation of $\alpha$ based on the variance of the point nearest-neighbour consistency loss from the previous generation. The variance in the nearest-neighbour consistency loss for different points in the background implies the distribution of dynamic objects in the scene. With fewer moving objects indicated by a lower variance, $\alpha$ should be adaptively reduced, tending to produce lower motion probability $P_M$ for points.

In addition to $L_{BG}$ and $L_{FG}$, we introduce box dimension regularization, heading term, and angle term to constrain the dimensions, heading, and rotation angles of the bounding boxes within a reasonable range \cite{deng2023rsf}. We also introduce a mass term to ensure that there are as many points as possible within the box, making the estimated motion parameters of the box more robust \cite{deng2023rsf}.

\subsection{Data Augmentation for 3D Flow Auto-labelling}
Existing data augmentation practices \cite{Pointpwc} often add consistent random rotations and noise offsets to the input points, which indeed yields certain benefits. However, in autonomous driving scenarios, there are frequently various complex motion patterns for multiple objects. To make models learn complex scene motion rules, we propose a novel data augmentation method for scene flow labelling in both global and object-level motions. Our method simulates a broad spectrum of 3D scene flow data variations, originating from ego-motion and dynamic object movement, thereby providing a promising solution to the challenge of securing abundant 3D scene flow labels.

\begin{figure}
  \centering
  \includegraphics[width=0.47\textwidth]{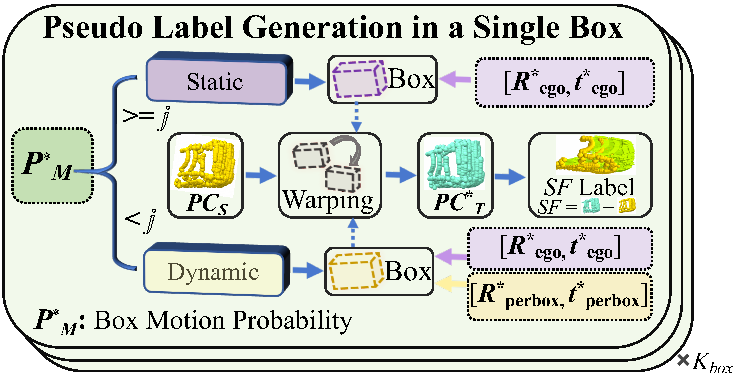}
  \vspace{-2mm}
  \caption{The proposed pseudo label generation module. With the augmented motion probability $P^*_M$, bounding boxes are categorized into dynamic and static types. Using global and local motion parameters, the $PC_S$ is warped to the target point cloud $PC^*_T$. Finally, pseudo 3D scene flow labels $SF$ are derived from the correspondence between $PC^*_T$ and $PC_S$. $K_{box}$ represents the number of boxes.}
  \label{fig:mo3}
\end{figure}
As in Fig. \ref{fig:method}, random noise is applied to either global or local motion parameters respectively. We generate a random rotation angle $\alpha$ and a random unit vector $\boldsymbol{u}$ for the rotation direction using random noise. They are used to create the Lie algebra $\boldsymbol{\xi}$. Subsequently, the Lie algebra $\boldsymbol{\xi}$ is converted into a rotation matrix $\mathbf{M}$ using the Rodrigues' rotation formula and applied to the original rotation matrix $\mathbf{R}$ to obtain a new rotation matrix $\mathbf{R}^*$, as follows:
\vspace{-3mm}
\begin{equation}
\label{eq:5}
\mathbf{M} = \mathbf{I} + \sin(|\boldsymbol{\xi}|)\frac{\boldsymbol{\xi}}{|\boldsymbol{\xi}|}_{\times} + (1 - \cos(|\boldsymbol{\xi}|))\left(\frac{\boldsymbol{\xi}}{|\boldsymbol{\xi}|}_{\times}\right)^2,
\end{equation}
\vspace{-2mm}
\begin{equation}
\label{eq:6}
\boldsymbol{\xi} = \alpha \boldsymbol{u}, \mathbf{R}^* = \mathbf{R} \mathbf{M}.
\end{equation}
The Lie algebra element $\boldsymbol{\xi}$, the product of scalar $\alpha$ and unit vector $\boldsymbol{u}$, signifies rotation magnitude and direction, with $\alpha$ and $\boldsymbol{u}$ representing rotation angle and axis, respectively. 
$\mathbf{I}$ is identity matrix, and $\boldsymbol{\xi}{\times}$ is the antisymmetric matrix of $\boldsymbol{\xi}$. Lie algebra intuitively and conveniently represents minor $SO(3)$ group variations. Rodrigues' rotation formula, mapping from the Lie algebra to the Lie group, facilitates the transformation of angle-based noise into a form directly applicable to the rotation matrix. This transformation brings mathematical convenience, making the update of the rotation matrix concise and efficient.

Importantly, our data augmentation targets dynamically moving objects, because persistently adding varied motion noise to bounding boxes perceived as static objects may disrupt original data distribution. Moreover, the translation and motion probability are also augmented. As depicted in Fig. \ref{fig:method}, we generate noise within an appropriate range and directly add it to the translation matrix or motion probability, resulting in augmented translation and motion probability.

\setlength{\tabcolsep}{0.6mm}
\begin{table*}[t]
	\begin{center}
		\caption{Comparison of our method with the best-performing methods on multiple datasets \cite{lidarKITTI,argoverse,nuscenes} and metrics. 'None', 'Weak', 'Self', and 'Full' represent non-learning, weakly supervised, self-supervised, and supervised methods, respectively. ``$\color{green}\bm{\uparrow}$'' means higher is better, and ``$\color{red}\bm{\downarrow}$'' means lower is better. Our method uses GMSF \cite{zhang2023gmsf} as a baseline and combines it with our proposed pseudo-auto-labelling framework, 3DSFlabelling. Despite the use of a supervised learning structure, no ground truth is utilized in training.}
		\label{table:1}
            \vspace{-2mm}
            \resizebox{1.0\textwidth}{!}{
		\begin{tabular}{cc|cccc|cccc|cccc}
			\toprule
			\multirow{2}{*}{Method} & \multirow{2}{*}{Sup.} & \multicolumn{4}{c}{LiDAR KITTI Scene Flow \cite{lidarKITTI}} & \multicolumn{4}{c}{Argoverse Scene Flow \cite{argoverse}} & \multicolumn{4}{c}{nuScenes Scene Flow \cite{nuscenes}} \\ 
		& & EPE3D$\color{red}\bm{\downarrow}$ & Acc3DS$\color{green}\bm{\uparrow}$  & Acc3DR$\color{green}\bm{\uparrow}$ &  Outliers$\color{red}\bm{\downarrow}$ & EPE3D$\color{red}\bm{\downarrow}$ & Acc3DS$\color{green}\bm{\uparrow}$  & Acc3DR$\color{green}\bm{\uparrow}$ &  Outliers$\color{red}\bm{\downarrow}$ & EPE3D$\color{red}\bm{\downarrow}$ & Acc3DS$\color{green}\bm{\uparrow}$  & Acc3DR$\color{green}\bm{\uparrow}$ &  Outliers$\color{red}\bm{\downarrow}$ \\  \midrule
                Graph prior \cite{Pontes} & None & -- & -- & -- & -- & 0.2570 & 0.2524 & 0.4760 & -- & 0.2890 & 0.2012 & 0.4354 & -- \\
                RSF \cite{deng2023rsf} & None & 0.0850 & 0.8830 & 0.9290 & 0.2390 & -- & -- & -- & -- & 0.1070 & 0.7170 & 0.8620 & 0.3210 \\
                NSFP \cite{li2021neural} & None & 0.1420 & 0.6880 & 0.8260 & 0.3850  & 0.1590 &  0.3843 &  0.6308 &  -- &  0.1751 & 0.3518 & 0.6345 & 0.5270 \\
                R3DSF \cite{gojcic2021weakly} & Weak & 0.0940 & 0.7840 & 0.8850 & 0.3140 & 0.4160 &  0.3452 & 0.4310 & 0.5580 & -- & -- & -- & -- \\ \midrule
                FlowNet3D \cite{Flownet3d} & Full & 0.7220 & 0.0300 & 0.1220 & 0.9650 & 0.4550 & 0.0134 & 0.0612 & 0.7360 & 0.5050 & 0.2120 & 0.1081 & 0.6200 \\
			PointPWC \cite{Pointpwc} & Full & 0.3900 & 0.3870 & 0.5500 & 0.6530 & 0.4288 & 0.0462 & 0.2164 & 0.9199 & 0.7883 & 0.0287 & 0.1333 & 0.9410 \\ 
                DCA-SRSFE \cite{jin2022deformation} & Full &  0.5900 &0.1505 &0.3331   &0.8485 &  0.7957 &0.0712 &0.1468  & 0.9799 &0.7042 & 0.0538 & 0.1183 & 0.9766 \\
			FLOT \cite{Flot} & Full & 0.6532 & 0.1554 &0.3130 &0.8371& 0.2491& 0.0946 & 0.3126 & 0.8657 & 0.4858&        0.0821& 0.2669 & 0.8547 \\ 
                MSBRN \cite{cheng2023multi} & Full & 0.0139 & 0.9752 & 0.9847  & 0.1433 & 0.8691 & 0.2432 & 0.2854   & 0.7597 & 0.6137 & 0.2354  & 0.2924 & 0.7638  \\ 
                GMSF \cite{zhang2023gmsf} & Full & 0.1900 & 0.2962 & 0.5502 & 0.6171 & 7.2776 & 0.0036 & 0.0144 & 0.9930 & 9.4231 & 0.0034 & 0.0086 & 0.9943\\ \midrule
			
            Mittal et al. \cite{Justgo} & Self & 0.9773 & 0.0096 & 0.0524 & 0.9936 & 0.6520 & 0.0319 & 0.1159 & 0.9621 & 0.8422 & 0.0289 &  0.1041 & 0.9615 \\ 
            Jiang et al. \cite{JiangPseudo} & Self & 0.4908 & 0.2052 & 0.4238 &  0.7286 & 0.2517 & 0.1236 & 0.3666 & 0.8114 & 0.4709 & 0.1034 & 0.3175 & 0.8191 \\
            Ours & Self & \bf0.0078 & \bf0.9924 & \bf0.9947 & \bf0.1328 & \bf0.0093 & \bf{0.9780} & \bf0.9880 & \bf0.1302 & \bf0.0185 & \bf0.9534 & \bf0.9713 & \bf0.1670 \\ 
			\bottomrule
		\end{tabular}
            }
	\end{center}
 \vspace{-2mm}
\end{table*}

\subsection{Pseudo Label Generation for 3D Scene Flow}
The motion parameters are fed into the pseudo label generation module to obtain point-wise 3D scene flow labels. The specific process of the label generation module is shown in Fig. \ref{fig:mo3}. We determine the motion state of the 3D bounding box through the motion probability $P_M$:
\begin{equation}
\label{eq:7}
PC^*_T = 
\begin{cases} 
PC_S \times R^*_{ego} + t^*_{ego}  & \text{if } P^*_M < \mathbb{J}, \\
PC_S^{ego} \times R^*_{perbox} + t^*_{perbox} & \text{if } P^*_M \geq \mathbb{J}.
\end{cases}
\end{equation}
$PC_S^{ego}$ is the points in the dynamic box from the source point cloud, transformed through global rotation and translation. 
When $P_M$ is less than threshold $\mathbb{J}$, the current bounding box is deemed static. Conversely, if $P_M$ exceeds a predefined threshold $\mathbb{J}$, the current bounding box is considered dynamic. For static boxes, based on the existing global motion, we apply a uniform noise to all static boxes to simulate various ego-motion patterns. By adding minute noise to the motion probability $P_M$ for each box, we can construct various motion states and show a greater variety of scene motions. Before transforming the dynamic boxes, a prior global transformation of all points is required. For dynamic bounding boxes, we add various noises to their existing motion, generating new rotations and translations, thereby creating various motion patterns. Once the motion state and parameters for each box are determined, we warp the source point cloud within each box to the target frame using the box's motion parameters, obtaining the pseudo target point cloud $PC^*_T$. 

The generated pseudo target point cloud $PC^*_T$ and the real source frame point cloud $PC_S$ have a perfect correspondence. Therefore, the 3D scene flow labels can be easily obtained by directly subtracting $PC_S$ from $PC^*_T$:
\begin{equation}
\label{eq:8}
SF = PC^*_T - PC_S.
\end{equation}
The generated scene flow labels capture various motion patterns from real autonomous driving scenes. They help the model understand and adjust to complex driving conditions. This improves the model's ability to generalize in unfamiliar real-world scenarios. 
\begin{figure*}[t]
  \centering
  \vspace{-2mm}
  \includegraphics[width=0.99\textwidth]{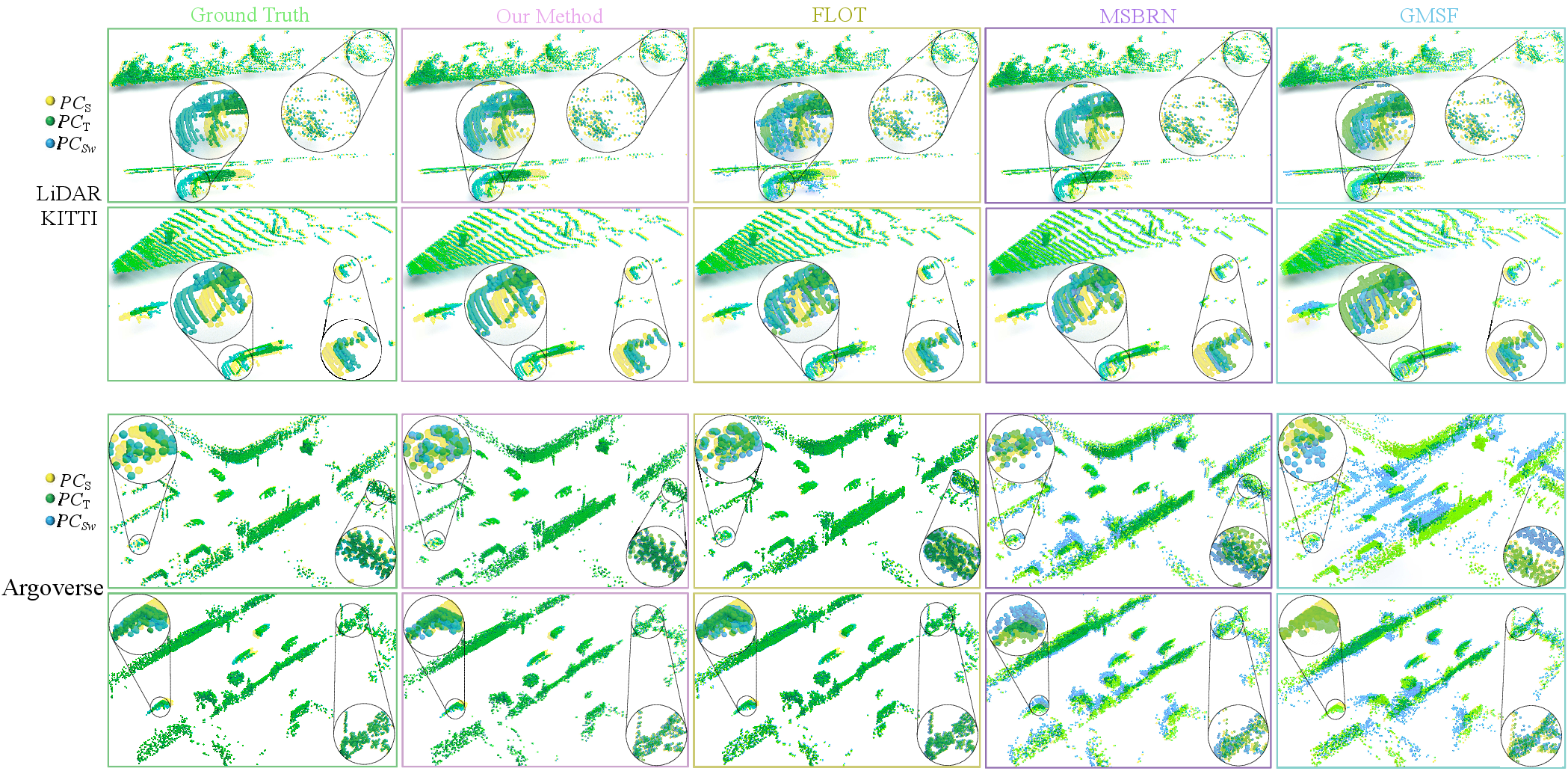}
  \vspace{-1mm}
  \caption{Registration visualization results of our method (GMSF \cite{zhang2023gmsf}+3DSFlabelling) and baselines on the LiDAR KITTI and Argoverse datasets \cite{lidarKITTI, argoverse}. The estimated target point cloud $PC_{sw}$ is derived from warping the source point cloud $PC_{S}$ to the target point cloud via 3D scene flow. The larger the overlap between $PC_{sw}$ (blue) and the target point cloud $PC_T$ (green), the higher the predicted accuracy of the scene flow. Local areas are zoomed in for better visibility. Our 3D scene flow estimation notably improves performance.
}
  \label{fig:vis1}

\end{figure*}

\section{Experiments} \label{Experiments}

\subsection{Datasets}
\textbf{Test Datasets:}  Graph prior \cite{Pontes} introduces two autonomous driving datasets, Argoverse scene flow \cite{argoverse} and nuScenes scene flow \cite{nuscenes} datasets. Scene flow labels in the datasets are derived from LiDAR point clouds, object trajectories, map data, and vehicle pose. The datasets contain 212 and 310 test samples, respectively. R3DSF \cite{gojcic2021weakly} introduces the lidarKITTI \cite{lidarKITTI}, which shares 142 scenes with stereoKITTI, collected via Velodyne's 64-beam LiDAR. Unlike FT3D \cite{FT3D} and stereoKITTI \cite{sfkitti,sfkitti2}, the point clouds from lidarKITTI are sparsely distributed. Note that LiDAR scene flow ground truths contain errors. We mitigate this by fusing the ground truth with the first point cloud to create a corrected second frame for network input, thus avoiding evaluation errors.

\begin{table}[t]
\begin{center}
\caption{The comparative results between our method and baseline. ``\textcolor{lightblue}{{\boldmath$\uparrow$}}'' signifies accuracy enhancement. In real-world LiDAR scenarios, our method markedly improves the 3D flow estimation accuracy across three datasets \cite{sfkitti,argoverse,nuscenes} on the three baselines. This demonstrates that the proposed pseudo-auto-labelling framework can substantially boost the accuracy of existing methods, even without the need for ground truth.}
\vspace{-2mm}
\label{table:22}
\resizebox{0.46\textwidth}{!}{
\begin{tabular}{ccccc}
\toprule
Dataset  & Method & EPE3D$\color{red}\bm{\downarrow}$ & Acc3DS$\color{green}\bm{\uparrow}$  & Acc3DR$\color{green}\bm{\uparrow}$ \\ \midrule

& \cellcolor[HTML]{C0C0C0}FLOT  \cite{Flot}& \multicolumn{1}{l}{\cellcolor[HTML]{C0C0C0}0.6532} & \cellcolor[HTML]{C0C0C0}0.1554                      & \cellcolor[HTML]{C0C0C0}0.3130                      \\
& FLOT+3DSFlabelling             & \textbf{0.0189} \textcolor{lightblue}{{\boldmath$\uparrow$} \textbf{97.1\%}}        & \textbf{0.9666}           & \textbf{0.9792}           \\
& \cellcolor[HTML]{C0C0C0}MSBRN \cite{cheng2023multi}& \multicolumn{1}{l}{\cellcolor[HTML]{C0C0C0}0.0139} & \cellcolor[HTML]{C0C0C0}0.9752                      & \cellcolor[HTML]{C0C0C0}0.9847                      \\
& MSBRN+3DSFlabelling            & \textbf{0.0123} \textcolor{lightblue}{{\boldmath$\uparrow$} \textbf{11.5\%}}         & \textbf{0.9797}   & \textbf{0.9868}   \\
& \cellcolor[HTML]{C0C0C0}GMSF \cite{zhang2023gmsf}  & \multicolumn{1}{l}{\cellcolor[HTML]{C0C0C0}0.1900} & \cellcolor[HTML]{C0C0C0}0.2962                      & \cellcolor[HTML]{C0C0C0}0.5502                      \\
\multirow{-6}{*}{\begin{tabular}[c]{@{}c@{}}LiDAR\\ KITTI\end{tabular}} & GMSF+3DSFlabelling             & \textbf{0.0078} \textcolor{lightblue}{{\boldmath$\uparrow$} \textbf{95.8\%}}        & \textbf{0.9924}   & \textbf{0.9947}   \\ \midrule
& \cellcolor[HTML]{C0C0C0}FLOT  \cite{Flot}& \multicolumn{1}{l}{\cellcolor[HTML]{C0C0C0}0.2491} & \cellcolor[HTML]{C0C0C0}0.0946                      & \cellcolor[HTML]{C0C0C0}0.3126                      \\
& FLOT+3DSFlabelling             & \textbf{0.0107} \textcolor{lightblue}{{\boldmath$\uparrow$} \textbf{95.7\%}}        & \textbf{0.9711}   & \textbf{0.9862}   \\
  Argoverse & \cellcolor[HTML]{C0C0C0}MSBRN \cite{cheng2023multi}&  \multicolumn{1}{l}{\cellcolor[HTML]{C0C0C0}0.8691} & \cellcolor[HTML]{C0C0C0}0.2432  & \cellcolor[HTML]{C0C0C0}0.2854       \\

 & MSBRN+3DSFlabelling            & \textbf{0.0150} \textcolor{lightblue}{{\boldmath$\uparrow$} \textbf{98.3\%}}         & \textbf{0.9482}   & \textbf{0.9601}   \\
& \cellcolor[HTML]{C0C0C0}GMSF \cite{zhang2023gmsf} & \multicolumn{1}{l}{\cellcolor[HTML]{C0C0C0}7.2776} & \cellcolor[HTML]{C0C0C0}0.0036                      & \cellcolor[HTML]{C0C0C0}0.0144                      \\
 & GMSF+3DSFlabelling  & \bf0.0093 \textcolor{lightblue}{{\boldmath$\uparrow$} \textbf{99.9\%}} & \bf{0.9780} & \bf0.9880 \\
\midrule
\multicolumn{1}{l}{}                & \cellcolor[HTML]{C0C0C0}FLOT  \cite{Flot}& \multicolumn{1}{l}{\cellcolor[HTML]{C0C0C0}0.4858} & \cellcolor[HTML]{C0C0C0}0.0821 & \cellcolor[HTML]{C0C0C0}0.2669 \\
\multicolumn{1}{l}{}                & FLOT+3DSFlabelling             & \textbf{0.0554} \textcolor{lightblue}{{\boldmath$\uparrow$} \textbf{88.6\%}}         & \textbf{0.7601}              & \textbf{0.8909}        \\
nuScenes & \cellcolor[HTML]{C0C0C0}MSBRN \cite{cheng2023multi}& \multicolumn{1}{l}{\cellcolor[HTML]{C0C0C0}0.6137} & \cellcolor[HTML]{C0C0C0}0.2354 & \cellcolor[HTML]{C0C0C0}0.2924 \\
\multicolumn{1}{l}{}                & MSBRN+3DSFlabelling            & \textbf{0.0235}  \textcolor{lightblue}{{\boldmath$\uparrow$} \textbf{96.2\%}}        & \textbf{0.9413}   & \textbf{0.9604}   \\
& \cellcolor[HTML]{C0C0C0}GMSF \cite{zhang2023gmsf}  & \multicolumn{1}{l}{\cellcolor[HTML]{C0C0C0}9.4231} & \cellcolor[HTML]{C0C0C0}0.0034                      & \cellcolor[HTML]{C0C0C0}0.0086                      \\
\multicolumn{1}{l}{}                       & GMSF+3DSFlabelling             & \textbf{0.0185} \textcolor{lightblue}{{\boldmath$\uparrow$} \textbf{99.8\%}}        & \textbf{0.9534}   & \textbf{0.9713} \\
\bottomrule
\end{tabular}
}
\end{center}
\vspace{-5mm}
\end{table}

\textbf{Training Datasets used in previous methods:} FT3D \cite{FT3D} and stereoKITTI \cite{sfkitti,sfkitti2} are the frequently used datasets for training previous 3D scene flow models \cite{Pointpwc,Flot,Flownet3d,cheng2023multi,zhang2023gmsf}. 
FT3D consists of 19,640 training pairs, while stereoKITTI \cite{sfkitti,sfkitti2} contains 142 dense point clouds, with the first 100 frames used for model fine-tuning in some works \cite{Justgo,li2021neural}. Some works \cite{Pontes,li2021neural,Flownet3d,Pointpwc,Justgo} train their models on 2,691 pairs of Argoverse \cite{argoverse} data and 1,513 pairs of nuScenes \cite{nuscenes} data, with 3D scene flow annotations following the settings of the Graph prior \cite{Pontes}. The R3DSF \cite{gojcic2021weakly} training set utilizes FT3D and semanticKITTI datasets \cite{behley2019semantickitti}, relying on ego-motion labels and semantic segmentation labels from semanticKITTI. 

\textbf{Training Datasets used in our methods:} Because we do not need any labels for training data, we use raw LiDAR point clouds sampled from raw data. For testing on the lidarKITTI \cite{sfkitti2}, we use LiDAR point clouds from sequences 00 to 09 of the KITTI Odometry dataset \cite{geiger2013vision} for auto-labelling and training. For testing on the nuScenes scene flow dataset \cite{nuscenes}, we randomly sample 50,000 pairs of LiDAR point clouds from the 350,000 LiDAR point clouds in the nuScenes sweeps dataset \cite{nuscenes}. For testing on the Argoverse scene flow Dataset \cite{argoverse}, we use the LiDAR point clouds from sequences 01 to 05 of the Argoverse 2 Sensor Dataset \cite{argoverse} for auto-labelling and training. In the selection of training data, we exclude the test scenes.

\subsection{Implementation Details}
The effectiveness of the proposed auto-labelling framework is demonstrated using three prominent deep learning models: FLOT \cite{Flot}, MSBRN \cite{cheng2023multi}, and GMSF \cite{zhang2023gmsf}. These models use optimal transport, coarse-to-fine strategies, and transformer architectures respectively. Hyperparameters consistent with the original networks are employed during the training process. The input point clouds, from which ground points have been filtered, are randomly sampled to incorporate 8192 points. The LiDAR point cloud data from KITTI \cite{lidarKITTI} is confined to the front view perspective, maintaining consistency with previous studies \cite{gojcic2021weakly}. Furthermore, we utilize four scene flow evaluation metrics \cite{Flownet3d, Flot, zhang2023gmsf, Pointpwc}: Average Endpoint Error (EPE3D), ACC3DS, ACC3DR, and Outliers. 


\setlength{\tabcolsep}{0.2mm}
\begin{table}[t]
	\begin{center}
	\caption{Model comparison on the Argoverse dataset \cite{argoverse}. 'M' represents millions of parameters, and time is in milliseconds.}
		\label{table:2}
            \vspace{-2mm}
            \resizebox{0.475\textwidth}{!}{
		\begin{tabular}{cc|ccc|cc}
			\toprule
			 Method & Sup.  & EPE3D$\color{red}\bm{\downarrow}$ & Acc3DS$\color{green}\bm{\uparrow}$  & Acc3DR$\color{green}\bm{\uparrow}$ &  Time$\color{red}\bm{\downarrow}$ & Params.$\color{red}\bm{\downarrow}$ \\  \midrule
            PointPWC \cite{Pointpwc} & Full & 0.4288 & 0.0462 & 0.2164 & 147 ms & 7.7 M \\
            PV‐RAFT \cite{PVRAFT} & Full & 10.745 & 0.0200 & 0.0100 & 169 ms & -- \\
            R3DSF \cite{gojcic2021weakly} & Weak & 0.4160 &  0.3452 & 0.4310 & 113 ms & 8.0 M \\
            FlowStep3D \cite{kittenplon2021flowstep3d} & Self &  0.8450 & 0.0100 & 0.0800 & 729 ms & -- \\
            NSFP \cite{li2021neural} & None & 0.1590 &  0.3843 &  0.6308 & 2864 ms & --  \\
            Fast-NSF \cite{FastNSF} & None & 0.1180 & 0.6993 & 0.8355 & 124 ms & --  \\
            MBNSF \cite{vidanapathirana2023multi} & None & 0.0510 & 0.7936 & 0.9237 & 5000+ ms & --\\
            MSBRN+3DSFlabelling & Self & 0.0150 & 0.9482 & 0.9601 & 341 ms & 3.5 M \\
            GMSF+3DSFlabelling & Self & \textbf{0.0093} & \textbf{0.9780} & \textbf{0.9880} & 251 ms & 6.0 M \\
            FLOT+3DSFlabelling & Self & 0.0107 & 0.9711 & 0.9862 & \textbf{78 ms} & \textbf{0.1 M} \\
			\bottomrule
		\end{tabular}
            }
	\end{center}
\vspace{-5mm}
\end{table}

\subsection{Quantitative Results}
The experimental results are presented in Table \ref{table:1}. We list the best-performing optimized \cite{Pontes,deng2023rsf,gojcic2021weakly,li2021neural}, self-supervised \cite{Justgo,JiangPseudo}, and supervised \cite{Flownet3d,Pointpwc,jin2022deformation} models in the table. Our method achieves excellent performance on all datasets \cite{argoverse,nuscenes,lidarKITTI} and metrics. Particularly, compared to the baselines \cite{zhang2023gmsf}, there is an order of magnitude reduction in EPE3D on most datasets. The proposed auto-labelling method generates effective scene flow labels, perfectly simulating the rigid motion of various objects in the real world. The designed global-local data augmentation further expands the 3D scene flow labels. As a result, our method significantly outperforms other methods. We have also applied this plug-and-play auto-labelling framework for 3D scene flow (3DSFlabelling) to three existing models, as demonstrated in Table \ref{table:22}. The proposed method significantly enhances the accuracy of 3D scene flow estimation in these models \cite{Flot,cheng2023multi,zhang2023gmsf}.

Moreover, many existing works utilize a large number of model parameters  \cite{Pointpwc,PVRAFT,gojcic2021weakly} or adopt optimization methods \cite{li2021neural,FastNSF,vidanapathirana2023multi} during testing for a more accurate estimation of 3D scene flow. These methods are highly time-consuming, and cannot ensure accuracy when reducing model parameters. Our proposed 3DSFlabelling effectively addresses this challenge. In Table \ref{table:2}, by using the small-parameter model FLOT (iter=1) \cite{Flot} combined with our auto-labelling framework, we surpass all current supervised, unsupervised, weakly supervised, and optimized methods. This strongly validates the effectiveness of generating real-world labels in solving the challenges.

\subsection{Visualization}
\begin{figure}[t]
  \centering
  \includegraphics[width=0.47\textwidth]{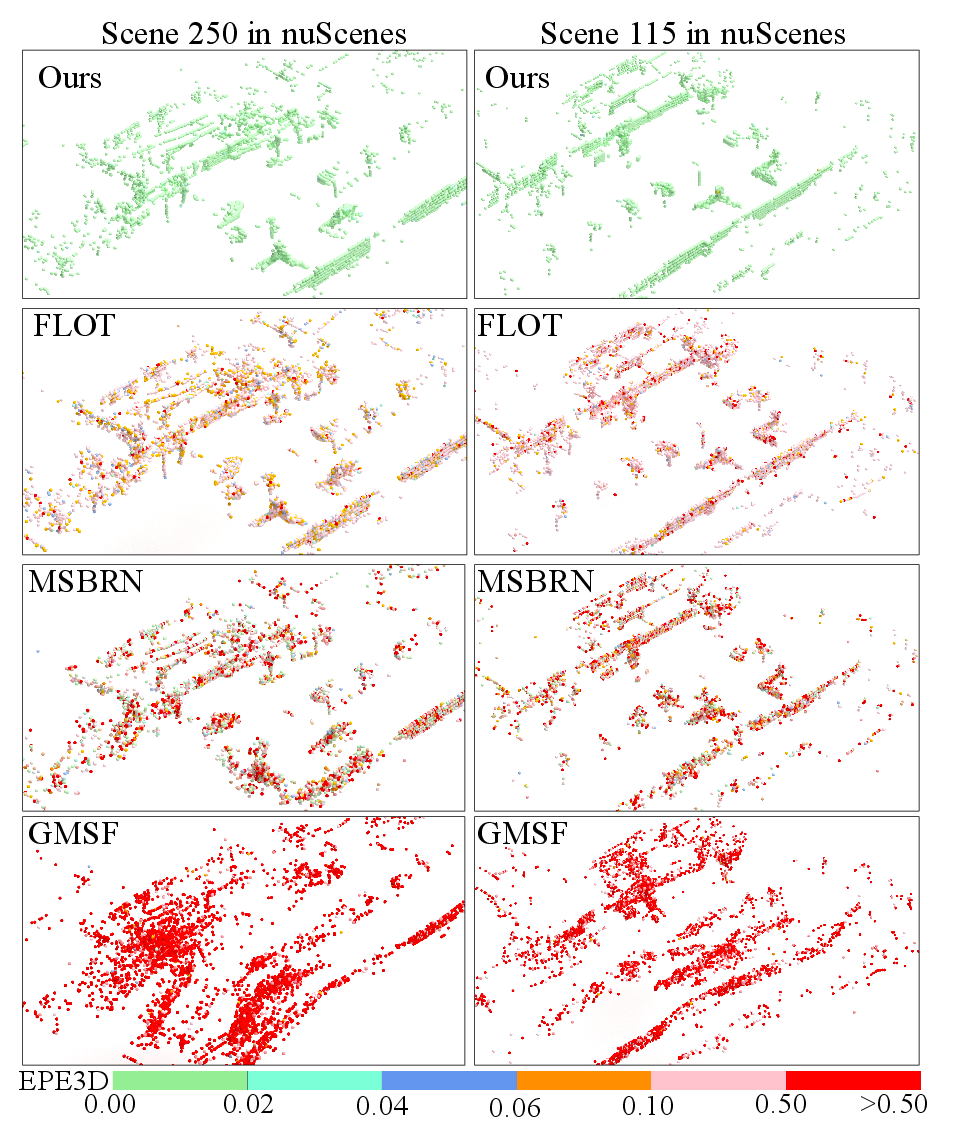}
  \vspace{-4mm}
  \caption{Error visualizing of our method (GMSF+3DSFlabelling) and baselines on the nuScenes dataset \cite{nuscenes}. Using 3D EndPoint Error (EPE3D) as the metric, we categorize the error into six levels. Combining GMSF \cite{zhang2023gmsf} with our proposed 3DSFlabelling, we manage to keep the EPE3D for most points within 0.02 meters, clearly outperforming other methods largely.}
  \label{fig:vis2}
\end{figure}

Fig. \ref{fig:vis1} visualizes the precision of our method and others on two datasets \cite{sfkitti2,argoverse}. FLOT \cite{Flot}, with its mathematically optimal transport approach to matching point clouds, exhibits superior generalization. MSBRN \cite{cheng2023multi}, leveraging a multi-scale bidirectional recurrent network, robustly estimates 3D scene flow on KITTI. GMSF \cite{zhang2023gmsf} utilizes a transformer architecture for powerful fitting learning, but it lacks cross-domain generalization. The proposed method consistently shows better alignment between predicted and target point clouds across all scenes. Additionally, a visualization of the scene flow error on the nuScenes dataset is presented in Fig. \ref{fig:vis2}. In two randomly selected test scenes, our method keeps the scene flow EPE3D mostly within 0.02$m$, clearly outperforming other baselines. More visual comparisons will be presented in the supplementary material.

\setlength{\tabcolsep}{0.1mm}
\begin{table}[t]
\centering
\caption{Generalization comparison experiment. ``A'', ``N'', and ``K'' represent the Argoverse \cite{argoverse}, nuScenes \cite{nuscenes}, and KITTI \cite{lidarKITTI} datasets. `$\rightsquigarrow$' representing a model trained on the dataset on the left and directly evaluated on another new dataset on the right.}
\label{table:3}
\vspace{-2mm}
\begin{scriptsize}
\renewcommand{\arraystretch}{1.2}
\begin{tabular}{cc|cc|cc|cc|cc}
\hline
\multirow{2}{*}{Method} & \multicolumn{2}{c}{\textbf{A$\rightsquigarrow$N}} & \multicolumn{2}{c}{\textbf{N$\rightsquigarrow$A}} & \multicolumn{2}{c}{\textbf{A$\rightsquigarrow$K}} & \multicolumn{2}{c}{\textbf{N$\rightsquigarrow$K}} \\ \cline{2-9} &  EPE3D & Acc3DS  & EPE3D & Acc3DS  & EPE3D & Acc3DS  & EPE3D & Acc3DS  \\ \hline
PointPWC \cite{Pointpwc} & 0.5911& 0.0844             & 0.7043& 0.0281             & 0.8632& 0.0119             & 0.9307& 0.0027             \\ \hline
RigidFlow \cite{gojcic2021weakly} & 0.1135& 0.3445             & 0.3991& 0.0152             & 0.3645& 0.2118 & 0.5042& 0.0141             \\ \hline
MSBRN \cite{cheng2023multi}& 0.5309 & 0.0055 & 0.3761 & 0.0098  & 0.6036 & 0.0056 & 0.4926 & 0.0081  \\ \hline
GMSF \cite{zhang2023gmsf}  & 0.0334& 0.9037             & 0.3078& 0.1278             & 0.0442& 0.8764 & 0.0574& 0.8135             \\ \hline
Ours   & \textbf{0.0115} & \textbf{0.9693} & \textbf{0.0264} & \textbf{0.9192}             & \textbf{0.0414} & \textbf{0.9020} & \textbf{0.0208} & \textbf{0.9595}    \\ 
\bottomrule
\end{tabular}
\end{scriptsize} 
\end{table}

Table \ref{table:3} provides quantitative results, demonstrating the generalization of our 3DSFlabelling combined with the existing method (GMSF \cite{zhang2023gmsf}) on new datasets. For instance, we train a model on the Argoverse dataset and directly evaluate it on the nuScenes dataset. These two datasets belong to different domains, posing a domain generalization problem. The results in Table \ref{table:3} indicate that our framework performs exceptionally well on the new dataset, consistently achieving an EPE3D of less than 5$cm$, and even reaching an average endpoint error of less than 2$cm$.
\subsection{Ablation Study}
This section explores the advantages of global-local data augmentation.
In Table \ref{table:4}, we compare existing 3D scene flow data augmentation \cite{Pointpwc} with our proposed global-local data augmentation method. Our augmentation strategy shows significant enhancement in all evaluation metrics. This is attributed to the effective simulation of various motion patterns in autonomous driving by global-local data augmentation. The introduction of various motion transformations excellently utilizes the limited training data to extend a variety of 3D scene flow styles. More ablation studies are referring to the supplement material.
\setlength{\tabcolsep}{0.2mm}
\begin{table}[t]
\centering
\caption{Ablation study of 3D scene flow data augmentation. ``No Aug'' and ``Trad. Aug'' represents no data augmentation and traditional data augmentation \cite{Pointpwc}, respectively. Our data augmentation method has a very positive impact on the model.}
\label{table:4}
    \vspace{-2mm}
    \resizebox{0.475\textwidth}{!}{
    \begin{tabular}{c|ccc|cc|cc|cc}
    \toprule
      \multirow{2}{*}{Model} & \multicolumn{3}{c|}{Data Augmentation Methods} & \multicolumn{2}{c|}{KITTI} & \multicolumn{2}{c|}{Argoverse} & \multicolumn{2}{c}{nuScenes} \\
 & No Aug   & Trad. Aug& Our Aug   & EPE3D & ACC3DS& EPE3D   & ACC3DS  & EPE3D   & ACC3DS \\ \midrule
     \multirow{3}{*}{\centering \begin{tabular}[t]{@{}m{1.5cm}@{}} Ours\\(FLOT)\end{tabular}} & \checkmark & -- & -- & 0.0601& 0.7291& 0.0492  & 0.8015  & 0.7364  & 0.6642 \\
       & -- & \checkmark & -- & 0.0540& 0.7622& 0.0430  & 0.8679  & 0.0610  & 0.7417 \\
          & -- & -- & \checkmark& \bf0.0189& \bf0.9666& \bf0.0107  & \bf0.9711  & \bf0.0554  & \bf0.7601 \\ \midrule

     \multirow{3}{*}{\centering \begin{tabular}[t]{@{}m{1.5cm}@{}} Ours\\(MSBRN)\end{tabular}} 
     & \checkmark & -- & -- & 0.0131 & 0.9781& 0.0180 & 0.9411  & 0.0797  & 0.8510 \\
      & -- & \checkmark & -- & 0.0129& 0.9790& 0.0177  & 0.9427  & 0.0793  & 0.8547 \\
          & -- & -- & \checkmark& \bf0.0123& \bf0.9797& \bf0.0150  & \bf0.9482  & \bf0.0235  & \bf0.9413 \\ \midrule
          
     \multirow{3}{*}{\centering \begin{tabular}[t]{@{}m{1.5cm}@{}} Ours\\(GMSF)\end{tabular}}
     & \checkmark & -- & -- & 0.0103& 0.9901& 0.0139  & 0.9637  & 0.0213  & 0.9468 \\
      & -- & \checkmark & -- & 0.0081& 0.9918& 0.0137  & 0.9663  & 0.0212  & 0.9473 \\
          & -- & -- & \checkmark& \bf0.0078& \bf0.9924& \bf0.0093  & \bf0.9780  & \bf0.0185  & \bf0.9534 \\ \bottomrule
\end{tabular}
}
\end{table}

\section{Conclusion}
We package 3D point clouds into boxes with different motion attributes. By optimizing the motion parameters for each box and warping the source point cloud into the target point cloud, we create pseudo 3D scene flow labels. We also design a global-local data augmentation method, introducing various scene motion patterns and significantly increasing the diversity and quantity of 3D scene flow labels. Tests on multiple real-world datasets show that our 3D scene flow auto-labelling significantly enhances the performance of existing models. Importantly, this approach eliminates the need for 3D scene flow estimation models to depend on manually annotated 3D scene flow labels.

{
    \small
    \bibliographystyle{ieeenat_fullname}
    \bibliography{main}

\begin{thebibliography}{48}
\providecommand{\natexlab}[1]{#1}
\providecommand{\url}[1]{\texttt{#1}}
\expandafter\ifx\csname urlstyle\endcsname\relax
  \providecommand{\doi}[1]{doi: #1}\else
  \providecommand{\doi}{doi: \begingroup \urlstyle{rm}\Url}\fi

\bibitem[Behley et~al.(2019)Behley, Garbade, Milioto, Quenzel, Behnke,
  Stachniss, and Gall]{behley2019semantickitti}
Jens Behley, Martin Garbade, Andres Milioto, Jan Quenzel, Sven Behnke, Cyrill
  Stachniss, and Jurgen Gall.
\newblock Semantickitti: A dataset for semantic scene understanding of lidar
  sequences.
\newblock In \emph{Proceedings of the IEEE/CVF international conference on
  computer vision}, pages 9297--9307, 2019.

\bibitem[Caesar et~al.(2020)Caesar, Bankiti, Lang, Vora, Liong, Xu, Krishnan,
  Pan, Baldan, and Beijbom]{nuscenes}
Holger Caesar, Varun Bankiti, Alex~H Lang, Sourabh Vora, Venice~Erin Liong,
  Qiang Xu, Anush Krishnan, Yu Pan, Giancarlo Baldan, and Oscar Beijbom.
\newblock nuscenes: A multimodal dataset for autonomous driving.
\newblock In \emph{Proceedings of the IEEE/CVF conference on computer vision
  and pattern recognition}, pages 11621--11631, 2020.

\bibitem[Chang et~al.(2019)Chang, Lambert, Sangkloy, Singh, Bak, Hartnett,
  Wang, Carr, Lucey, Ramanan, et~al.]{argoverse}
Ming-Fang Chang, John Lambert, Patsorn Sangkloy, Jagjeet Singh, Slawomir Bak,
  Andrew Hartnett, De Wang, Peter Carr, Simon Lucey, Deva Ramanan, et~al.
\newblock Argoverse: 3d tracking and forecasting with rich maps.
\newblock In \emph{Proceedings of the IEEE/CVF Conference on Computer Vision
  and Pattern Recognition}, pages 8748--8757, 2019.

\bibitem[Cheng and Ko(2023)]{cheng2023multi}
Wencan Cheng and Jong~Hwan Ko.
\newblock Multi-scale bidirectional recurrent network with hybrid correlation
  for point cloud based scene flow estimation.
\newblock In \emph{Proceedings of the IEEE/CVF International Conference on
  Computer Vision}, pages 10041--10050, 2023.

\bibitem[Deng and Zakhor(2023)]{deng2023rsf}
David Deng and Avideh Zakhor.
\newblock Rsf: Optimizing rigid scene flow from 3d point clouds without labels.
\newblock In \emph{Proceedings of the IEEE/CVF Winter Conference on
  Applications of Computer Vision}, pages 1277--1286, 2023.

\bibitem[Ding et~al.(2022)Ding, Pan, Deng, Deng, and Lu]{ding2022self}
Fangqiang Ding, Zhijun Pan, Yimin Deng, Jianning Deng, and Chris~Xiaoxuan Lu.
\newblock Self-supervised scene flow estimation with 4-d automotive radar.
\newblock \emph{IEEE Robotics and Automation Letters}, 7\penalty0 (3):\penalty0
  8233--8240, 2022.

\bibitem[Dong et~al.(2022)Dong, Zhang, Li, Sun, and Xiong]{dong2022exploiting}
Guanting Dong, Yueyi Zhang, Hanlin Li, Xiaoyan Sun, and Zhiwei Xiong.
\newblock Exploiting rigidity constraints for lidar scene flow estimation.
\newblock In \emph{Proceedings of the IEEE/CVF Conference on Computer Vision
  and Pattern Recognition}, pages 12776--12785, 2022.

\bibitem[Er{\c{c}}elik et~al.(2022)Er{\c{c}}elik, Yurtsever, Liu, Yang, Zhang,
  Top{\c{c}}am, Listl, Cayl{\i}, and Knoll]{erccelik20223d}
Eme{\c{c}} Er{\c{c}}elik, Ekim Yurtsever, Mingyu Liu, Zhijie Yang, Hanzhen
  Zhang, P{\i}nar Top{\c{c}}am, Maximilian Listl, Y{\i}lmaz~Kaan Cayl{\i}, and
  Alois Knoll.
\newblock 3d object detection with a self-supervised lidar scene flow backbone.
\newblock In \emph{European Conference on Computer Vision}, pages 247--265.
  Springer, 2022.

\bibitem[Fu et~al.(2023)Fu, Xiang, Qiao, and Bai]{fu2023pt}
Jingyun Fu, Zhiyu Xiang, Chengyu Qiao, and Tingming Bai.
\newblock Pt-flownet: Scene flow estimation on point clouds with point
  transformer.
\newblock \emph{IEEE Robotics and Automation Letters}, 8\penalty0 (5):\penalty0
  2566--2573, 2023.

\bibitem[Geiger et~al.(2012)Geiger, Lenz, and Urtasun]{lidarKITTI}
Andreas Geiger, Philip Lenz, and Raquel Urtasun.
\newblock Are we ready for autonomous driving? the kitti vision benchmark
  suite.
\newblock In \emph{2012 IEEE conference on computer vision and pattern
  recognition}, pages 3354--3361. IEEE, 2012.

\bibitem[Geiger et~al.(2013)Geiger, Lenz, Stiller, and
  Urtasun]{geiger2013vision}
Andreas Geiger, Philip Lenz, Christoph Stiller, and Raquel Urtasun.
\newblock Vision meets robotics: The kitti dataset.
\newblock \emph{The International Journal of Robotics Research}, 32\penalty0
  (11):\penalty0 1231--1237, 2013.

\bibitem[Gojcic et~al.(2021)Gojcic, Litany, Wieser, Guibas, and
  Birdal]{gojcic2021weakly}
Zan Gojcic, Or Litany, Andreas Wieser, Leonidas~J Guibas, and Tolga Birdal.
\newblock Weakly supervised learning of rigid 3d scene flow.
\newblock In \emph{Proceedings of the IEEE/CVF conference on computer vision
  and pattern recognition}, pages 5692--5703, 2021.

\bibitem[Gu et~al.(2019)Gu, Wang, Wu, Lee, and Wang]{Hplflownet}
Xiuye Gu, Yijie Wang, Chongruo Wu, Yong~Jae Lee, and Panqu Wang.
\newblock Hplflownet: Hierarchical permutohedral lattice flownet for scene flow
  estimation on large-scale point clouds.
\newblock In \emph{Proceedings of the IEEE/CVF Conference on Computer Vision
  and Pattern Recognition}, pages 3254--3263, 2019.

\bibitem[Huang et~al.(2022)Huang, Gojcic, Huang, Wieser, and
  Schindler]{huang2022dynamic}
Shengyu Huang, Zan Gojcic, Jiahui Huang, Andreas Wieser, and Konrad Schindler.
\newblock Dynamic 3d scene analysis by point cloud accumulation.
\newblock In \emph{European Conference on Computer Vision}, pages 674--690.
  Springer, 2022.

\bibitem[Iqbal et~al.(2021)Iqbal, Al-Kaff, Marin, Marcenaro, Gomez, and
  Regazzoni]{iqbal2021detection}
Hafsa Iqbal, Abdulla Al-Kaff, Pablo Marin, Lucio Marcenaro, David~Martin Gomez,
  and Carlo Regazzoni.
\newblock Detection of abnormal motion by estimating scene flows of point
  clouds for autonomous driving.
\newblock In \emph{2021 IEEE International Intelligent Transportation Systems
  Conference (ITSC)}, pages 2788--2793. IEEE, 2021.

\bibitem[Jiang et~al.(2023)Jiang, Wang, Miao, and Wang]{JiangPseudo}
Chaokang Jiang, Guangming Wang, Yanzi Miao, and Hesheng Wang.
\newblock 3-d scene flow estimation on pseudo-lidar: Bridging the gap on
  estimating point motion.
\newblock \emph{IEEE Transactions on Industrial Informatics}, 19\penalty0
  (6):\penalty0 7346--7354, 2023.

\bibitem[Jin et~al.(2022)Jin, Lei, Akhtar, Li, and Hayat]{jin2022deformation}
Zhao Jin, Yinjie Lei, Naveed Akhtar, Haifeng Li, and Munawar Hayat.
\newblock Deformation and correspondence aware unsupervised synthetic-to-real
  scene flow estimation for point clouds.
\newblock In \emph{Proceedings of the IEEE/CVF Conference on Computer Vision
  and Pattern Recognition}, pages 7233--7243, 2022.

\bibitem[Kittenplon et~al.(2021)Kittenplon, Eldar, and
  Raviv]{kittenplon2021flowstep3d}
Yair Kittenplon, Yonina~C Eldar, and Dan Raviv.
\newblock Flowstep3d: Model unrolling for self-supervised scene flow
  estimation.
\newblock In \emph{Proceedings of the IEEE/CVF Conference on Computer Vision
  and Pattern Recognition}, pages 4114--4123, 2021.

\bibitem[Lang et~al.(2023)Lang, Aiger, Cole, Avidan, and
  Rubinstein]{lang2023scoop}
Itai Lang, Dror Aiger, Forrester Cole, Shai Avidan, and Michael Rubinstein.
\newblock Scoop: Self-supervised correspondence and optimization-based scene
  flow.
\newblock In \emph{Proceedings of the IEEE/CVF Conference on Computer Vision
  and Pattern Recognition}, pages 5281--5290, 2023.

\bibitem[Li et~al.(2022{\natexlab{a}})Li, Zheng, Giancola, and
  Ghanem]{li2022sctn}
Bing Li, Cheng Zheng, Silvio Giancola, and Bernard Ghanem.
\newblock Sctn: Sparse convolution-transformer network for scene flow
  estimation.
\newblock In \emph{Proceedings of the AAAI Conference on Artificial
  Intelligence}, pages 1254--1262, 2022{\natexlab{a}}.

\bibitem[Li et~al.(2022{\natexlab{b}})Li, Zhang, Lin, Wang, and
  Shen]{li2022rigidflow}
Ruibo Li, Chi Zhang, Guosheng Lin, Zhe Wang, and Chunhua Shen.
\newblock Rigidflow: Self-supervised scene flow learning on point clouds by
  local rigidity prior.
\newblock In \emph{Proceedings of the IEEE/CVF Conference on Computer Vision
  and Pattern Recognition}, pages 16959--16968, 2022{\natexlab{b}}.

\bibitem[Li et~al.(2021)Li, Kaesemodel~Pontes, and Lucey]{li2021neural}
Xueqian Li, Jhony Kaesemodel~Pontes, and Simon Lucey.
\newblock Neural scene flow prior.
\newblock \emph{Advances in Neural Information Processing Systems},
  34:\penalty0 7838--7851, 2021.

\bibitem[Li et~al.(2023)Li, Zheng, Ferroni, Pontes, and Lucey]{FastNSF}
Xueqian Li, Jianqiao Zheng, Francesco Ferroni, Jhony~Kaesemodel Pontes, and
  Simon Lucey.
\newblock Fast neural scene flow.
\newblock In \emph{Proceedings of the IEEE/CVF International Conference on
  Computer Vision (ICCV)}, pages 9878--9890, 2023.

\bibitem[Liu et~al.(2019)Liu, Qi, and Guibas]{Flownet3d}
Xingyu Liu, Charles~R Qi, and Leonidas~J Guibas.
\newblock Flownet3d: Learning scene flow in 3d point clouds.
\newblock In \emph{Proceedings of the IEEE/CVF Conference on Computer Vision
  and Pattern Recognition}, pages 529--537, 2019.

\bibitem[Mayer et~al.(2016)Mayer, Ilg, Hausser, Fischer, Cremers, Dosovitskiy,
  and Brox]{FT3D}
Nikolaus Mayer, Eddy Ilg, Philip Hausser, Philipp Fischer, Daniel Cremers,
  Alexey Dosovitskiy, and Thomas Brox.
\newblock A large dataset to train convolutional networks for disparity,
  optical flow, and scene flow estimation.
\newblock In \emph{Proceedings of the IEEE Conference on Computer Vision and
  Pattern Recognition}, pages 4040--4048, 2016.

\bibitem[Menze et~al.(2015)Menze, Heipke, and Geiger]{sfkitti}
Moritz Menze, Christian Heipke, and Andreas Geiger.
\newblock Joint 3d estimation of vehicles and scene flow.
\newblock \emph{ISPRS annals of the photogrammetry, remote sensing and spatial
  information sciences}, 2:\penalty0 427--434, 2015.

\bibitem[Menze et~al.(2018)Menze, Heipke, and Geiger]{sfkitti2}
Moritz Menze, Christian Heipke, and Andreas Geiger.
\newblock Object scene flow.
\newblock \emph{ISPRS Journal of Photogrammetry and Remote Sensing},
  140:\penalty0 60--76, 2018.

\bibitem[Mittal et~al.(2020)Mittal, Okorn, and Held]{Justgo}
Himangi Mittal, Brian Okorn, and David Held.
\newblock Just go with the flow: Self-supervised scene flow estimation.
\newblock In \emph{Proceedings of the IEEE/CVF Conference on Computer Vision
  and Pattern Recognition}, pages 11177--11185, 2020.

\bibitem[Najibi et~al.(2022)Najibi, Ji, Zhou, Qi, Yan, Ettinger, and
  Anguelov]{najibi2022motion}
Mahyar Najibi, Jingwei Ji, Yin Zhou, Charles~R Qi, Xinchen Yan, Scott Ettinger,
  and Dragomir Anguelov.
\newblock Motion inspired unsupervised perception and prediction in autonomous
  driving.
\newblock In \emph{European Conference on Computer Vision}, pages 424--443.
  Springer, 2022.

\bibitem[Peng et~al.(2023)Peng, Wang, Lo, Wu, Xu, Tomizuka, Zhan, and
  Wang]{peng2023delflow}
Chensheng Peng, Guangming Wang, Xian~Wan Lo, Xinrui Wu, Chenfeng Xu, Masayoshi
  Tomizuka, Wei Zhan, and Hesheng Wang.
\newblock Delflow: Dense efficient learning of scene flow for large-scale point
  clouds.
\newblock In \emph{Proceedings of the IEEE/CVF International Conference on
  Computer Vision}, pages 16901--16910, 2023.

\bibitem[Pontes et~al.(2020)Pontes, Hays, and Lucey]{Pontes}
Jhony~Kaesemodel Pontes, James Hays, and Simon Lucey.
\newblock Scene flow from point clouds with or without learning.
\newblock In \emph{2020 International Conference on 3D Vision (3DV)}, pages
  261--270, 2020.

\bibitem[Puy et~al.(2020)Puy, Boulch, and Marlet]{Flot}
Gilles Puy, Alexandre Boulch, and Renaud Marlet.
\newblock Flot: Scene flow on point clouds guided by optimal transport.
\newblock In \emph{ECCV 2020: 16th European Conference, Glasgow, UK, August
  23--28, 2020, Proceedings, Part XXVIII 16}, pages 527--544, 2020.

\bibitem[Qi et~al.(2017{\natexlab{a}})Qi, Su, Mo, and Guibas]{qi2017pointnet}
Charles~R Qi, Hao Su, Kaichun Mo, and Leonidas~J Guibas.
\newblock Pointnet: Deep learning on point sets for 3d classification and
  segmentation.
\newblock In \emph{Proceedings of the IEEE conference on computer vision and
  pattern recognition}, pages 652--660, 2017{\natexlab{a}}.

\bibitem[Qi et~al.(2017{\natexlab{b}})Qi, Yi, Su, and Guibas]{qi2017pointnet++}
Charles~Ruizhongtai Qi, Li Yi, Hao Su, and Leonidas~J Guibas.
\newblock Pointnet++: Deep hierarchical feature learning on point sets in a
  metric space.
\newblock \emph{Advances in neural information processing systems}, 30,
  2017{\natexlab{b}}.

\bibitem[Shen et~al.(2023)Shen, Hui, Xie, and Yang]{Shen_2023_CVPR}
Yaqi Shen, Le Hui, Jin Xie, and Jian Yang.
\newblock Self-supervised 3d scene flow estimation guided by superpoints.
\newblock In \emph{Proceedings of the IEEE/CVF Conference on Computer Vision
  and Pattern Recognition (CVPR)}, pages 5271--5280, 2023.

\bibitem[Tishchenko et~al.(2020)Tishchenko, Lombardi, Oswald, and
  Pollefeys]{tishchenko2020self}
Ivan Tishchenko, Sandro Lombardi, Martin~R Oswald, and Marc Pollefeys.
\newblock Self-supervised learning of non-rigid residual flow and ego-motion.
\newblock In \emph{2020 international conference on 3D vision (3DV)}, pages
  150--159. IEEE, 2020.

\bibitem[Vidanapathirana et~al.(2023)Vidanapathirana, Chng, Li, and
  Lucey]{vidanapathirana2023multi}
Kavisha Vidanapathirana, Shin-Fang Chng, Xueqian Li, and Simon Lucey.
\newblock Multi-body neural scene flow.
\newblock \emph{arXiv preprint arXiv:2310.10301}, 2023.

\bibitem[Wang et~al.(2021)Wang, Wu, Liu, and Wang]{wang2021hierarchical}
Guangming Wang, Xinrui Wu, Zhe Liu, and Hesheng Wang.
\newblock Hierarchical attention learning of scene flow in 3d point clouds.
\newblock \emph{IEEE Transactions on Image Processing}, 30:\penalty0
  5168--5181, 2021.

\bibitem[Wang et~al.(2022{\natexlab{a}})Wang, Hu, Liu, Zhou, Tomizuka, Zhan,
  and Wang]{wang2022matters}
Guangming Wang, Yunzhe Hu, Zhe Liu, Yiyang Zhou, Masayoshi Tomizuka, Wei Zhan,
  and Hesheng Wang.
\newblock What matters for 3d scene flow network.
\newblock In \emph{European Conference on Computer Vision}, pages 38--55.
  Springer, 2022{\natexlab{a}}.

\bibitem[Wang et~al.(2022{\natexlab{b}})Wang, Jiang, Shen, Miao, and
  Wang]{wang2022sfgan}
Guangming Wang, Chaokang Jiang, Zehang Shen, Yanzi Miao, and Hesheng Wang.
\newblock Sfgan: Unsupervised generative adversarial learning of 3d scene flow
  from the 3d scene self.
\newblock \emph{Advanced Intelligent Systems}, 4\penalty0 (4):\penalty0
  2100197, 2022{\natexlab{b}}.

\bibitem[Wang et~al.(2023{\natexlab{a}})Wang, Chi, and
  Yang]{wang2023exploiting}
Yun Wang, Cheng Chi, and Xin Yang.
\newblock Exploiting implicit rigidity constraints via weight-sharing
  aggregation for scene flow estimation from point clouds.
\newblock \emph{arXiv preprint arXiv:2303.02454}, 2023{\natexlab{a}}.

\bibitem[Wang et~al.(2023{\natexlab{b}})Wang, Wei, Rao, Zhou, and Lu]{DPVRAFT}
Ziyi Wang, Yi Wei, Yongming Rao, Jie Zhou, and Jiwen Lu.
\newblock 3d point-voxel correlation fields for scene flow estimation.
\newblock \emph{IEEE Transactions on Pattern Analysis and Machine
  Intelligence}, 2023{\natexlab{b}}.

\bibitem[Wei et~al.(2021)Wei, Wang, Rao, Lu, and Zhou]{PVRAFT}
Yi Wei, Ziyi Wang, Yongming Rao, Jiwen Lu, and Jie Zhou.
\newblock Pv-raft: Point-voxel correlation fields for scene flow estimation of
  point clouds.
\newblock In \emph{Proceedings of the IEEE/CVF Conference on Computer Vision
  and Pattern Recognition (CVPR)}, pages 6954--6963, 2021.

\bibitem[Wu et~al.(2020{\natexlab{a}})Wu, Chen, and Metaxas]{wu2020motionnet}
Pengxiang Wu, Siheng Chen, and Dimitris~N Metaxas.
\newblock Motionnet: Joint perception and motion prediction for autonomous
  driving based on bird's eye view maps.
\newblock In \emph{Proceedings of the IEEE/CVF conference on computer vision
  and pattern recognition}, pages 11385--11395, 2020{\natexlab{a}}.

\bibitem[Wu et~al.(2020{\natexlab{b}})Wu, Wang, Li, Liu, and Fuxin]{Pointpwc}
Wenxuan Wu, Zhi~Yuan Wang, Zhuwen Li, Wei Liu, and Li Fuxin.
\newblock Pointpwc-net: Cost volume on point clouds for (self-) supervised
  scene flow estimation.
\newblock In \emph{European Conference on Computer Vision}, pages 88--107,
  2020{\natexlab{b}}.

\bibitem[Zhang et~al.(2020)Zhang, Ye, Xiang, and Gu]{zhang2020sdp}
Yi Zhang, Yuwen Ye, Zhiyu Xiang, and Jiaqi Gu.
\newblock Sdp-net: Scene flow based real-time object detection and prediction
  from sequential 3d point clouds.
\newblock In \emph{Proceedings of the Asian Conference on Computer Vision},
  2020.

\bibitem[Zhang et~al.(2023)Zhang, Edstedt, Wandt, Forss{\'e}n, Magnusson, and
  Felsberg]{zhang2023gmsf}
Yushan Zhang, Johan Edstedt, Bastian Wandt, Per-Erik Forss{\'e}n, Maria
  Magnusson, and Michael Felsberg.
\newblock Gmsf: Global matching scene flow.
\newblock \emph{arXiv preprint arXiv:2305.17432}, 2023.

\bibitem[Zhou et~al.(2018)Zhou, Park, and Koltun]{zhou2018open3d}
Qian-Yi Zhou, Jaesik Park, and Vladlen Koltun.
\newblock Open3d: A modern library for 3d data processing.
\newblock \emph{arXiv preprint arXiv:1801.09847}, 2018.

\end{thebibliography}
}


\end{document}